%% file: iclr2026_conference.tex
\documentclass{article} % For LaTeX2e
\usepackage{iclr2026_conference,times}

\input{math_commands.tex}

\usepackage{booktabs}
\usepackage{hyperref}
\usepackage{url}
\usepackage{graphicx}
\usepackage{wrapfig}
\usepackage{multirow}
\usepackage{tcolorbox}
\usepackage{listings}
\usepackage{enumitem} % For more control over lists
\usepackage{amssymb} % For \boxed
\usepackage{array} % 用于更高级的列格式

\tcbuselibrary{breakable, skins}

\lstdefinestyle{jsonstyle_black}{
    language=json,
    basicstyle=\ttfamily\small,
    showstringspaces=false,
    breaklines=true,
    keywordstyle=\bfseries,
    stringstyle=\itshape,
    literate=
      *{:}{{{\bfseries:}}}1
       {,}{{{,}}}1
       {\{}{{{\bfseries\{}}}1
       {\}}{{{\bfseries\}}}}1
       {[}{{{\bfseries[}}}1
       {]}{{{\bfseries]}}}1,
}

\lstdefinestyle{promptstyle}{
    basicstyle=\ttfamily,
    breaklines=true,
    keepspaces=true,
    showstringspaces=false
}

\lstdefinestyle{tablestyle}{
    backgroundcolor=\color{white},
    basicstyle=\small\ttfamily,
    breaklines=true,
}

\title{GeoSketch: A Neural-Symbolic Approach to Geometric Multimodal Reasoning with Auxiliary Line Construction and Affine Transformation}

\author{
Shichao Weng $ ^{1} $\,\thanks{Work done during an internship at Hainan University.} \\
\texttt{scweng23@m.fudan.edu.cn}\\
Fudan University, Shanghai, China\\
\And Zhiqiang Wang  $ ^{1} $ \\
\texttt{zqwang42@iflytek.com} \\
IFLYTEK  CO.LTD \\
\And Yuhua Zhou $ ^{1} $ \\
\texttt{zhouyuhua@zju.edu.cn} \\
Zhejiang University, Zhejiang, China \\
\And Rui Lu \\
\texttt{rluac@connect.ust.hk} \\
The Hong Kong University of Science and Technology
\And Ting Liu \\
\texttt{liuting20@nudt.edu.cn} \\
National University of Defense Technology, Hunan, China
\And Zhiyang Teng \\
\texttt{chihyangteng@gmail.com} \\
 \\
\And Xiaozhang Liu \\
\texttt{lxzh@hainanu.edu.cn} \\
Hainan University, Haikou, China \\
\And Hanmeng Liu \thanks{Corresponding author.}\\
\texttt{liuhanmeng@hainanu.edu.cn} \\
Hainan University, Haikou, China \\
}

\iclrfinalcopy % Uncomment for camera-ready version, but NOT for submission.
\begin{document}

\maketitle

\begin{abstract}
Geometric Problem Solving (GPS) poses a unique challenge for Multimodal Large Language Models (MLLMs), requiring not only the joint interpretation of text and diagrams but also iterative visuospatial reasoning. While existing approaches process diagrams as static images, they lack the capacity for dynamic manipulation—a core aspect of human geometric reasoning involving auxiliary line construction and affine transformations.
We present GeoSketch, a neural-symbolic framework that recasts geometric reasoning as an interactive perception–reasoning–action loop. GeoSketch integrates: (1) a Perception module that abstracts diagrams into structured logic forms, (2) a Symbolic Reasoning module that applies geometric theorems to decide the next deductive step, and (3) a Sketch Action module that executes operations such as drawing auxiliary lines or applying transformations, thereby updating the diagram in a closed loop. To train this agent, we develop a two-stage pipeline: supervised fine-tuning on 2,000 symbolic-curated trajectories followed by reinforcement learning with dense, symbolic rewards to enhance robustness and strategic exploration.
To evaluate this paradigm, we introduce the GeoSketch Benchmark, a high-quality set of 390 geometry problems requiring auxiliary construction or affine transformations. Experiments on strong MLLM baselines demonstrate that GeoSketch significantly improves stepwise reasoning accuracy and problem-solving success over static perception methods.
By unifying hierarchical decision-making, executable visual actions, and symbolic verification, GeoSketch advances multimodal reasoning from static interpretation to dynamic, verifiable interaction, establishing a new foundation for solving complex visuospatial problems.
\end{abstract}

\section{Introduction}

With the advent of Multimodal Large Language Models (MLLMs) \citep{openai2024gpt4ocard,comanici2025gemini,hong2025glm}, Geometric Problem Solving (GPS) presents a unique challenge to MLLMs, demanding not only the joint understanding of text and diagrams but also rigorous, multi-step deductive reasoning \citep{10.24963/ijcai.2023/376,qiao2024wemathdoeslargemultimodal,he2025matpbenchmllmgoodautomated}. 
While modern MLLMs can ingest multimodal inputs, their reasoning output remains confined to static text.  
This limits the use of dynamic and visuospatial strategies in geometric problem solving, which becomes particularly evident in complex tasks requiring multi-stage manipulation.

\begin{wrapfigure}{r}{0.7\textwidth}
 \centering
  \includegraphics[width=0.7\textwidth]{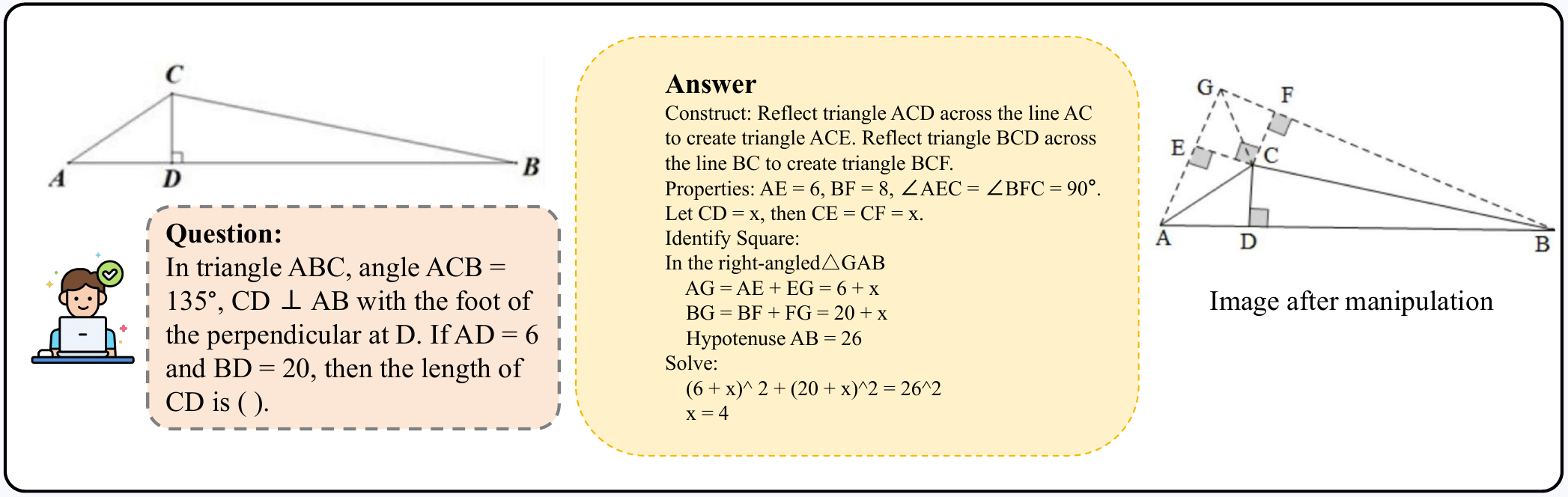}
  \caption{An visual geometric problem that require dynamic image manipulation}
  \label{fig:teaser}
  \vspace{-10pt}
\end{wrapfigure}

Human geometric problem-solving is inherently interactive and dynamic \citep{christou2005problem,freksa2019geometric}. As shown in Figure \ref{fig:teaser}, confronted with a complex proof, experts do not merely analyze a static diagram; they actively manipulate it through a temporal sequence of actions. This process involves auxiliary line construction to reveal latent properties and affine transformations—including reflection, rotation, and translation—to reconfigure components, establish congruences, and deduce new relationships. Each action is a deliberate step in a logical argument, creating a visual proof that evolves over time. This synergy of symbolic logic and spatial manipulation remains a critical gap in AI capabilities \citep{10.1145/3711680,wang2025multimodal,yan2025surveymathematicalreasoningera,cho2025planegeometryproblemsolving}.

%\begin{figure}
%    \centering
%    \includegraphics[width=0.7\linewidth]{iclr2026/figures/teaser.pdf}
%    \caption{An visual geometric problem that require dynamic image manipulation}
%    %\vspace{-2em}
%    \label{fig:teaser}
%\end{figure}

As the field of multimodal reasoning evolving, it has been moving beyond static visual perception towards a ``think with image'' cognitive workspace \citep{su2025thinkingimagesmultimodalreasoning}. In GPS, the challenge lies in encoding this step-by-step reasoning process that mirrors how humans intuitively manipulate geometric objects to solve problems.
Recent advancements, such as Visual Sketchpad \citep{10.5555/3737916.3742339}, have introduced image-based reasoning through sketching. 
Yet, these approaches predominantly treat sketching as a monolithic final answer, failing to model the executable, iterative reasoning process that leads to it. True geometric reasoning requires a dynamic workspace where logical thought and spatial action are intertwined.

To bridge this gap, we introduce GeoSketch, a neuro-symbolic framework that recasts geometric problem-solving as a structured, multimodal interaction loop. GeoSketch enables dynamic, ordered manipulation of the diagram, allowing an agent to incrementally build a solution through a synergy of symbolic logic and visual execution. Our framework is built on a hierarchical three-tier architecture that mirrors the human problem-solving process:
1) \textbf{Visual Perception}: Processes the input diagram using visual tools (e.g., YOLO \citep{Redmon2015YouOL}) to abstract it into a structured, symbolic representation (logic forms). This step distills the pixel-based image into a set of objects and relations that can be rendered losslessly, forming a manipulable digital twin of the diagram.
2) \textbf{Symbolic Reasoning}: A theorem-proving module, powered by an instructed MLLM, operates on this symbolic representation. It applies geometric knowledge to deduce the next logical step and decides on a specific, executable sketch operation (e.g., ``draw a line from point A to point B" or ``rotate triangle ABC by 90 degrees").
3) \textbf{Sketch Action}: A rendering engine executes the commanded operation, altering the diagram. This updated visual state is then fed back into the perception module, closing the loop and enabling the next cycle of perception, reasoning, and action.
This architecture ensures that every visual manipulation is grounded in symbolic logic, mitigating hallucination and fostering verifiable reasoning.

To instantiate GeoSketch, we implement it on the Qwen2.5-VL-7B foundation model and adopt a two-stage training strategy. First, we use supervised fine-tuning (SFT) on a high-quality, curated dataset of 2,000 expert trajectories. These trajectories are generated using geometric templates and handcrafted rules to guarantee correctness and diversity, teaching the model valid step-by-step reasoning. Second, we employ reinforcement learning (RL) to allow the model to explore the vast space of valid reasoning paths, enhancing its strategic flexibility and robustness against novel problems.
To support and evaluate this new paradigm, we contribute two key resources: first is the 2K GeoSketch fine-tuning dataset, a distilled collection of multi-step visual reasoning trajectories, and second, the GeoSketch benchmark, the largest curated set of 390 geometric problems specifically designed to require auxiliary construction and affine transformations for their solution. Experiment results show that the GeoSketch benchmark challenges current MLLMs, and our architecture boosts the performance compared to former methods.

By unifying hierarchical decision-making, differentiable rendering, and symbolic verification, GeoSketch advances multimodal reasoning from static perception to dynamic, executable interaction. It establishes a new foundation for tackling complex visuospatial problems where the process is as important as the answer.

\section{Related Work}

\subsection{Automated Geometry Theorem Proving and Problem Solving}
Automated geometry theorem proving has evolved from symbolic deduction to multi-modal reasoning. Early systems like iGeoTutor \citep{10.5555/2832249.2832414} and Inter-GPS \citep{lu-etal-2021-inter} translate diagrams into formal logic and apply symbolic rules to generate human-readable proofs. By reasoning with standard Euclidean axioms rather than complex algebraic proofs, these symbolic systems produce interpretable and verifiable solutions. The symbolic components of GeoSketch follow this line of research as the proofs are both friendly to humans and LLMs.

A significant challenge in this domain is the scarcity of high-quality, structured training data. GeoDRL \citep{peng2023geodrl} addressed this by coupling symbolic theorem solvers with a Graph Neural Network (GNN) trained via self-supervised reinforcement learning on geometry logic graphs. Google’s Olympiad Solver \citep{Trinh2024SolvingOG} pushed this paradigm further by generating massive-scale synthetic data to train a specialized transformer model from scratch, enabling it to generate complete proofs conditioned on theorem premises and conclusions. In contrast to these specialized architectures, GeoSketch adopts an agentic training framework built upon a general-purpose Multimodal Large Language Model (MLLM), leveraging its inherent reasoning and multimodal capabilities.
UniGeo \citep{chen-etal-2022-unigeo} unifies the problem of calculation and proving in the form of sequence generation. Similarly, GeoSketch also unifies different forms of the testing data (multi-choice, gap filling, and proofing) into a gold-labelled answer generation task.
Recent work has focused on leveraging large-scale vision-language fine-tuning.
GeoGPT4V \citep{cai-etal-2024-geogpt4v} synthesizes basic geometry problems with aligned text and images by leveraging GPT-4 and GPT-4V. Similarly, G-LLAVA \citep{gao2025gllava} is trained on 170K geometric image-caption and question-answer pairs to outperform GPT4-V on the geometric task of MathVista \citep{lu2024mathvista} benchmark. While these models address visual question answering, they operate on a static image representation and lack the capability to dynamically manipulate the diagram, which is a core contribution of GeoSketch.
Although neural-symbolic mechanisms have been explored by GeoDRL and Google's Olympiad solver, less work has been done on MLLMs.
AutoGPS \citep{10.5555/3737916.3742339} introduces a neuro-symbolic collaborative framework to solve geometry problems with MLLMs. It uses diagram and text parsers, along with a deductive symbolic reasoner, to overcome the deficiency of MLLMs. Similar to their method, GeoSketch uses tools to help with image perception and reasoning. Our critical advancement beyond AutoGPS and all prior work is the introduction of an executable action space. GeoSketch does not just reason about the diagram; it reasons with the diagram by generating a sequence of image manipulation actions (e.g., drawing auxiliary lines, performing transformations), thereby closing the loop between symbolic thought and visuospatial execution.

\subsection{Edit-Based Visual Reasoning}

While nascent in geometric reasoning, the paradigm of dynamically editing images during the reasoning process has emerged as a powerful approach for complex multimodal tasks. 

\paragraph{General-Purpose Visual Editing} \citep{fang2025got} re-conceptualizes text-to-image generation as an interactive, reasoning-guided editing process, allowing users to refine outputs by modifying intermediate reasoning steps. However, its reliance on diffusion models makes it unsuitable for the precise, structured manipulations required in geometric deduction. Similarly, Draw with Thought (DwT) \citep{cui2025drawthoughtunleashingmultimodal} introduces a training-free framework for editing scientific diagrams by reconstructing their mxGraph XML representations. By treating the diagram as code, it leverages the code-generation capabilities of MLLMs. GeoSketch adopts a similar principle of using an intermediate representation but utilizes declarative logic forms instead of imperative XML code. This offers superior clarity, readability, and direct alignment with symbolic theorem provers.

\paragraph{Sequential Sketching and Interaction}
The importance of temporal sequencing is recognized in SketchAgent \citep{vinker2024sketchagentlanguagedrivensequentialsketch}, which enables users to create and refine sketches conversationally through a sequence of strokes. While it captures the dynamic nature of sketching, its focus on free-form vector graphics for drawing is distinct from GeoSketch's goal of executing precise, theorem-driven geometric constructions. The SKETCHPAD \citep{10.5555/3737916.3742339} framework and its tutoring extension, INTERACTIVE SKETCHPAD \citep{chen2025interactivesketchpadmultimodaltutoring}, prompt Vision-Language Models (VLMs) to produce sketches as part of a chain-of-thought. Although evaluated on a couple of geometry problems, their action space is typically limited to simple line drawing and the interaction is primarily between a human and an LLM for tutoring. GeoSketch advances this by formalizing sketching as a temporal process of executable actions (including transformations), where the agent itself interacts with an evolving diagram to solve a problem autonomously.

\paragraph{Benchmarks for Edit-Based Reasoning}
Several benchmarks have been developed to assess dynamic visual reasoning. RBench-V \citep{guo2025rbenchv} provides a standardized framework for evaluating multi-modal reasoning, while CoMT \citep{10.1609/aaai.v39i22.34538} incorporates visual creation, deletion, and update into chain-of-thought benchmarking. Others, like VisAidMath \citep{ma2024visaidmathbenchmarkingvisualaidedmathematical}, MV-MATH \citep{wang2025mv}, and VisuLogic \citep{xu2025visulogic}, integrate visual contexts into mathematical reasoning. A key limitation of these benchmarks is that they are not specifically designed for geometric reasoning; only a small fraction of their problems involve geometry, and crucially, none evaluate the temporal evolution of a solution sketch. The GeoSketch benchmark directly addresses this gap by providing a curated set of problems that require multi-step diagram manipulation as an integral part of the reasoning process.

\section{Dataset and Methodology}

\begin{figure}
    \centering
    \includegraphics[width=0.9\linewidth]{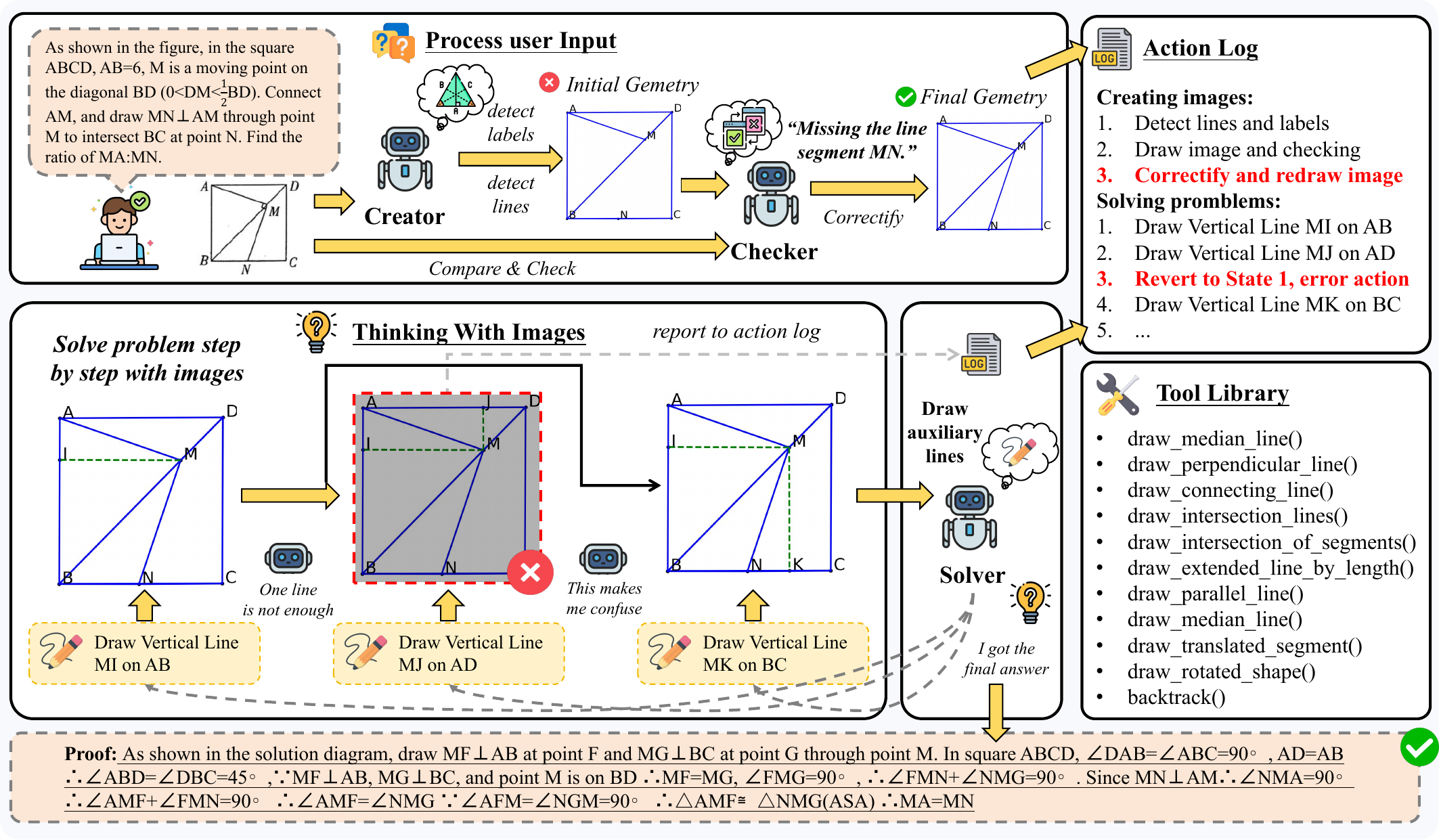}
    \caption{The three-tier GeoSketch architecture first uses its perception module to get the initial logic form and then refine the logic form until it gets a complete one. Then the neural-symbolic reasoning module tries to solve the question by drawing auxiliary lines, and the sketch-action module manipulates the image.}
    \label{fig:overview}
    \vspace{-10pt}
\end{figure}

\subsection{The GeoSketch Benchmark and Fine-Tuning Dataset}

The GeoSketch benchmark is curated from a corpus of approximately 1,200 geometry problems. To focus on dynamic visual reasoning, we apply a rigorous multi-stage filtering process, selecting only problems that require auxiliary line construction or affine transformations (reflection, rotation, translation). Expert annotators verify solution accuracy and diagram clarity, and the final problems are translated into English and standardized into a definitive question-answer format. Questions originally in multiple-choice, fill-in-the-blank, or proof-based formats are rewritten into a standardized format with definitive gold answers; problems that cannot be transformed decisively are discarded. Each finalized problem is represented as a structured JSON object containing a unique identifier, a clear textual description, paths to the initial and solution diagram images, the final gold answer, and a detailed step-by-step natural language proof. This process results in a high-quality benchmark of 390 problems. The data statistics and format are detailed in Appendix~\ref{app:statistics}.

The GeoSketch fine-tuning dataset is generated via a neural-symbolic pipeline designed to produce a large volume of high-quality, verifiable reasoning trajectories, with an emphasis on ``hard mode'' training scenarios. This approach systematically synthesizes complex geometric problems through template-based generation and rule-based difficulty enhancement, ensuring the model encounters challenging multi-step reasoning tasks.
Following \citet{lu-etal-2021-inter}, the pipeline begins with a seed set of geometric problem templates incorporating advanced theorems and complex configurations. These templates are instantiated using handcrafted rules crafted to maximize difficulty by introducing non-obvious auxiliary line requirements, multi-layered transformation sequences, and constrained geometric conditions. To boost diversity, we employ multiple advanced language models (\textsc{Gemini-2.5-pro}, \textsc{Qwen3} \citep{yang2025qwen3technicalreport}, \textsc{Seed1.6-Thinking} \citep{seed2025seed1}, and \textsc{DeepSeek-R1} \citep{deepseekai2025deepseekr1incentivizingreasoningcapability}) collaboratively during the generation process.
The formal logic representations are rendered into PNG images using a matplotlib-based agent, with an auto-correction mechanism applied to ensure coordinate precision. Each generated problem undergoes human-supervised difficulty scoring based on proof length and theorem complexity, resulting in a challenging fine-tuning dataset that emphasizes hard-mode training.

\subsection{The GeoSketch Neuro-Symbolic Architecture}

The GeoSketch architecture is designed as a hierarchical, three-tier agent that operates in a closed loop, iteratively refining a geometric diagram to arrive at a solution. As illustrated in Figure \ref{fig:overview}, the core cycle consists of a Perception module that parses the visual workspace, a Neural-Symbolic Reasoning module that orchestrates LLMs with logic forms, and a Sketch Action module that executes the decision, thereby altering the workspace for the next iteration. This design formalizes the dynamic interplay between visual perception and symbolic deduction that characterizes expert human problem-solving.

\subsubsection{Perception Module}

The Perception Module is responsible for converting the visual state of the diagram into a structured, symbolic representation amenable to logical manipulation. We employ a fine-tuned YOLOv11 \citep{khanam2024yolov11overviewkeyarchitectural} object detection model, a fine-tuned U-net \citep{DBLP:journals/corr/RonnebergerFB15} segment model and an OCR model to identify and classify primitive geometric elements within the diagram. The output of this module is a structured logic form—a JSON-based representation that enumerates all geometric objects, their properties, and their relations. A typical entry may define a line segment by its endpoints {type: `line', points: [`A', `B']} or a circle by its center and radius {type: `circle', center: `O', radius: 5}. This symbolic representation is both human-readable and machine-executable, and it can be rendered back into a standard visual diagram losslessly, ensuring a tight coupling between the visual and symbolic states.

One key improvement of our perception module is the auto error correction mechanism. As frameworks like Visual Sketchpad rely on manual corrections on each problem to obtain the logic form, we solved this key drawback by leveraging MLLMs to recognize the differences between the original diagram and the one generated using the logic form. As depicted in Figure \ref{fig:overview}, the MLLMs will find the mistakes in the logic form obtained from the perception module and refine the logic form till the generated diagram matches the original one which eliminates the requirements for human inspection.

\subsubsection{Neural-Symbolic Reasoning Module}

At the heart of the agent is the Neural-Symbolic Reasoning Module, which is implemented by instructing multimodal language models to function as a geometric theorem prover. Its input is a structured prompt containing the natural language problem text, the current logic form representing the diagram state, and the history of previous reasoning steps and actions. The module is tasked with generating the next step in the proof. Its output is constrained to a specific JSON format containing two critical fields: a natural language ``reasoning" field that justifies the step, and an ``action" field containing a command for the sketch module. We evaluate popular MLLMs as the backbones, as well as a smaller Qwen2.5-VL-7B model, through a combination of supervised fine-tuning on our curated dataset of expert trajectories and an RL process design that primes the model to handle complex geometric deductive principles.

\subsubsection{Sketch Action Module}

The Sketch Action Module translates the symbolic decisions from the reasoner into concrete manipulations of the diagram. Its functionality is defined by a precise action space. The primary executable actions include: draw\_line(point\_i, point\_j) to construct an auxiliary line; reflect(object, line\_of\_symmetry), rotate(object, center, degrees), and translate(object, vector) for affine transformations; and label\_point(name, coordinates) for annotating new elements. An action executor engine, which operates on the logic form, interprets these commands. For instance, upon receiving a rotate command, the executor updates the coordinates of all points within the specified object in the logic form and triggers a re-rendering of the visual diagram. This update completes the agent's loop, as the newly altered diagram becomes the input for the next perception step.

\subsection{Training Methodology}
To train a compact Qwen2.5VL-7B model to rival the performance of larger, more advanced MLLMs, we propose a two-stage strategy. This approach combines supervised learning through knowledge distillation to acquire foundational skills, followed by reinforcement learning to enhance strategic reasoning and robustness.

\subsubsection{Supervised Fine-Tuning (SFT) via Knowledge Distillation}

The initial training stage employs Supervised Fine-Tuning (SFT) to instill the model with core competencies in geometric deduction and structured action generation. To ensure the highest quality training data, we adopt a knowledge distillation approach within a teacher-student paradigm.

In this setup, the larger and more capable MLLMs act as the ``teacher''. We leverage this teacher models with our neural-symbolic pipeline to generate rigorous step-by-step expert trajectories for the problems in our fine-tuning dataset, as described in Section \ref{sec:sft-data}. Our data-generation pipeline enables MLLMs to generate ``hard-to-find'' questions rather than just distilling common questions. Each trajectory generated by the teacher is formatted as a multi-turn conversation. At each turn, the input consists of the history of previous auxiliary line drawing operations and the new geometric elements generated by these constructions. The target output, which serves as the ground truth for our smaller ``student" model (Qwen2.5VL-7B), is the teacher's structured response—a natural language reasoning step and a JSON object containing the corresponding executable action command.
This SFT process effectively transfers the nuanced geometric reasoning and precise structured-output capabilities from the large teacher model to our more efficient student model. By conditioning the training on the history of the reasoning process, this method enables our model to learn the sequential and conditional nature of complex geometric problem-solving.

\subsubsection{Reinforcement Learning (RL)}

While SFT provides a strong foundation, it can lead to brittle behavior where the model struggles to recover from errors or explore alternative valid paths. To address this, we employ a second Reinforcement Learning (RL) stage to enhance the model's strategic flexibility and robustness by allowing it to explore the vast state-action space of geometric reasoning.
In this framework, the GeoSketch architecture serves as the environment. The state is the current logic form of the diagram, and an action is the JSON command generated by the model. The agent's policy is initialized from the SFT-trained Qwen2.5VL-7B model.

To optimize the model's policy, we utilize the Group Relative Policy Optimization (GRPO)\citep{shao2024deepseekmath} algorithm, which has proven effective for incentivizing the inherent reasoning capability of MLLMs. 

A critical component of our RL stage is a sparse, rule-based reward function designed to directly optimize for task success. The reward is composed of two binary components awarded at the end of each trajectory:

\begin{itemize}
    \item \textbf{Format Reward ($R_{\text{format}}$):} A reward of +1.0 is given if the model's final output strictly adheres to the predefined JSON schema for reasoning and actions. Otherwise, the reward is 0. This encourages the model to generate syntactically valid and executable commands.
    \item \textbf{Result Reward ($R_{\text{result}}$):} A reward of +1.0 is given if the model's final computed answer matches the gold-standard solution for the problem. Otherwise, the reward is 0. This directly optimizes for problem-solving accuracy.
\end{itemize}

This RL stage encourages the agent to not only find correct solutions but also to consistently produce valid outputs, significantly improving its generalization capabilities and reliability on novel and complex problems.

\section{Experiment}

\subsection{Experiment Setup}

To evaluate the efficacy of the GeoSketch framework, we conduct a comprehensive set of experiments on our newly curated GeoSketch benchmark. Given that prior approaches like Visual Sketchpad cannot autonomously solve these problems without human intervention, our experimental design focuses on rigorously comparing the performance of leading Multimodal Large Language Models (MLLMs) under the following conditions:

\paragraph{Direct Evaluation (Pure Run)} In this baseline setting, we assess the models' inherent geometric reasoning capabilities in a static, single-turn inference mode. Each MLLM is provided with a standard prompt containing the problem's text and its corresponding initial diagram image. The model is then tasked with generating the final answer directly. This condition represents the standard paradigm for MLLM evaluation and measures the limits of static multimodal perception and reasoning.

\paragraph{GeoSketch (Dynamic Interaction)} In this experimental condition, we integrate the same MLLMs as the neural backbone of the symbolic reasoning module within our GeoSketch architecture. Instead of a single inference step, the model engages in the iterative perception–reasoning–action loop. At each step, it receives the current diagram state as a logic form and the reasoning history, and its task is to generate the next logical step and an executable sketch action. The loop continues until the model produces a terminal answer or reaches a predefined step limit. This setup evaluates the performance gains unlocked by dynamic diagram manipulation and structured symbolic reasoning.

\paragraph{Models and Metrics}
We evaluate a diverse set of state-of-the-art MLLMs, including the proprietary \textsc{GPT-4o} and powerful open-source models like \textsc{Qwen2.5VL-72B} and \textsc{Qwen2.5VL-7B}. Furthermore, we assess our two custom-trained models, \textbf{GeoSketch-Qwen2.5VL-7B-SFT} and \textbf{GeoSketch-Qwen2.5VL-7B-RL}, to demonstrate the effectiveness of our training methodology. Human performance is given by the average score of 3 testees.

The primary metric is final answer accuracy, defined as the percentage of problems for which a model provides the correct gold-standard answer. To provide a more granular analysis, we categorize problems by their answer type—\textbf{Numerical} (requiring a numeric value), \textbf{Ratio} (requiring a fraction or ratio), and \textbf{Descriptor} (requiring a textual proof or logical conclusion)—and report accuracy for each sub-category.

\subsection{Results and Analysis}

\begin{table}[t]
\caption{Performance comparison of MLLMs on the GeoSketch benchmark. The ``Pure Run" column shows the accuracy of models operating statically on images, while the ``GeoSketch" column shows the accuracy when models are empowered with dynamic reasoning and action capabilities. The ``Improvement" rows highlight the substantial gains achieved with our framework. Our final RL-tuned model achieves state-of-the-art performance, outperforming even significantly larger models. All results are in accuracy \%.}
\label{tab:main_results}
\centering
\resizebox{\textwidth}{!}{
\begin{tabular}{l|cccc|cccc}
\toprule
\multirow{2}{*}{\textbf{Model}} & \multicolumn{4}{c|}{\textbf{Pure Run (Baseline)}} & \multicolumn{4}{c}{\textbf{GeoSketch Framework}} \\
\cmidrule(lr){2-5} \cmidrule(lr){6-9}
& \textbf{Total} & \textbf{Numerical} & \textbf{Ratio} & \textbf{Descriptor} & \textbf{Total} & \textbf{Numerical} & \textbf{Ratio} & \textbf{Descriptor} \\
\midrule
\textbf{GPT-4o} & 33.08 & 24.88 & 50.00 & 30.86 & 49.74 & 44.28 & 59.26 & 50.62 \\
\textit{\quad Improvement} & & & & & \textit{+16.66} & \textit{+19.40} & \textit{+9.26} & \textit{+19.76} \\
\midrule
\textbf{Qwen2.5VL-72B} & 43.08 & 38.81 & 49.07 & 45.68 & 58.71 & 57.71 & 56.48 & 64.20 \\
\textit{\quad Improvement} & & & & & \textit{+15.63} & \textit{+18.90} & \textit{+7.41} & \textit{+18.52} \\
\textbf{Qwen2.5VL-7B} & 32.56 & 23.88 & 50.00 & 30.86 & 48.46 & 45.77 & 57.40 & 43.21 \\
\textit{\quad Improvement} & & & & & \textit{+15.90} & \textit{+21.89} & \textit{+7.40} & \textit{+12.35} \\
\midrule
\textbf{GeoSketch-Qwen2.5VL-7B-SFT} & 33.85 & 25.37 & 50.93 & 32.10 & 52.56 & 49.75 & 60.19 & 49.38 \\
\textit{\quad Improvement} & & & & & \textit{+18.71} & \textit{+24.38} & \textit{+9.26} & \textit{+17.28} \\
\textbf{GeoSketch-Qwen2.5VL-7B-RL} & 35.64 & 28.36 & 48.15 & 37.04 & 59.23 & 57.21 & 60.19 & 62.96 \\
\textit{\quad Improvement} & & & & & \textit{+23.59} & \textit{+28.85} & \textit{+12.04} & \textit{+25.92} \\
\midrule
\textbf{Human} & 86.14 & 84.08 & 87.04 & 90.12 & - & - & - & - \\
\bottomrule
\end{tabular}}
\end{table}

Experiment results are presented in Table \ref{tab:main_results}. 
The results demonstrate that GeoSketch poses significant challenges to MLLMs, highlighting the difficulty of this task in model processing. The best performance model in Pure Run, Qwen2.5VL-72B, give an accuracy of 43.08 \%, far below human average.
Across every model tested, integrating it within the GeoSketch framework yields an evident improvement in geometric problem-solving accuracy. Even the most powerful models, \textsc{GPT-4o} and \textsc{Qwen2.5VL-72B}, see their performance leap by +16.66 and +15.63 points, respectively. This demonstrates that raw model scale is insufficient to overcome the inherent limitations of static reasoning. The interactive loop, which allows models to build solutions incrementally and visually verify each deductive step, is a critical missing component that our framework provides. The largest gains are consistently observed in numerical and descriptive problems, where multi-step calculations and logical constructions are most prevalent.

%\paragraph{Two-Stage Training Strategy Delivers SOTA Performance}
Our core contribution is further validated by the exceptional performance of our custom-trained models. The SFT-trained model, \textbf{GeoSketch-Qwen2.5VL-7B-SFT}, already surpasses its base model by a significant margin (+4.1 points in the GeoSketch setting), confirming the value of distilling knowledge from high-quality expert trajectories.
The subsequent RL stage elevates performance to a new level. Our final model, \textbf{GeoSketch-Qwen2.5VL-7B-RL}, not only achieves the highest overall accuracy of \textbf{59.23\%} but also records the largest improvement (+23.59 points) over its pure-run baseline. This result highlights the effectiveness of RL in refining strategic decision-making and enhancing robustness. By exploring the vast reasoning space and receiving targeted rewards for correctness, the agent learns more effective and reliable problem-solving policies.
Our experiment results indicate that a specialized agent architecture is able to outperforms larger generalists. Our 7B parameter model, after undergoing the full SFT and RL training pipeline, marginally outperforms the much larger \textsc{Qwen2.5VL-72B} (59.23\% vs. 58.71\%) and substantially surpasses \textsc{GPT-4o} (59.23\% vs. 49.74\%) within the GeoSketch framework. This demonstrates that a smaller, specialized agent equipped with the right interactive tools and trained with a targeted methodology can achieve superior performance on complex, domain-specific tasks compared to larger, general-purpose models.
Overall, by recasting geometric problem-solving from a static interpretation task to a dynamic, interactive process, the GeoSketch framework provides the necessary tools for this paradigm shift, and our two-stage training strategy effectively cultivates the specialized reasoning skills required to master it.

\section{Discussion}

Our experiments have demonstrated the substantial performance gains achieved by integrating MLLMs into the GeoSketch framework for complex geometric problems requiring dynamic manipulation. In this section, we further explore the broader implications of our approach by examining its generalization capabilities on standard geometry benchmarks and comparing its operational efficiency against related methods.

\subsection{Generalization to Standard Geometric Problems}
A key question is whether training an agent for the specialized task of dynamic interaction also enhances its fundamental geometric reasoning abilities on static problems. To investigate this, we evaluated our fine-tuned models on two established benchmarks, Geometry3K \citep{lu-etal-2021-inter} and GeoQA \citep{DBLP:journals/corr/abs-2105-14517}, which consist of general geometry problems that do not necessarily require auxiliary line construction or affine transformations. The base Qwen2.5VL-7B model serves as our baseline.

\begin{table}[h]
\caption{Performance on general geometry benchmarks (GEO3K and GEOQA). The results show that the GeoSketch training pipeline significantly improves the model's performance on standard, static geometry problems, indicating an enhancement of core geometric knowledge. All results are in accuracy \%.}
\label{tab:generalization_results}
\centering
\scalebox{0.9}{
\begin{tabular}{lcc}
\toprule
\textbf{Model} & \textbf{Geometry3K} & \textbf{GeoQA} \\
\midrule
\textbf{Qwen2.5VL-7B (Base)} & 19.30 & 47.28 \\
\textbf{GeoSketch-Qwen2.5VL-7B-SFT}         & 22.46 & 65.09 \\
\textbf{GeoSketch-Qwen2.5VL-7B-RL}          & \textbf{28.79} & \textbf{72.52} \\
\bottomrule
\end{tabular}
}
\end{table}

As shown in Table \ref{tab:generalization_results}, both the SFT and RL stages of our training pipeline yield consistent improvements over the baseline model. On Geometry3K, the RL-tuned model achieves an accuracy of 28.79\%, a 9.49 percentage point improvement over the base model. The gains are even more pronounced on GeoQA, where the RL model reaches 72.52\% accuracy, a remarkable 25.24 point increase.

These results strongly suggest that our training methodology does more than just teach tool use. The high-quality, structured trajectories in our fine-tuning dataset effectively instill fundamental geometric knowledge and deductive reasoning patterns into the model. The SFT stage provides a solid foundation, and the subsequent RL stage refines this understanding by encouraging exploration and reinforcing successful problem-solving strategies. Consequently, even when the model is not provided with the interactive GeoSketch loop, its intrinsic ability to reason about geometric configurations is significantly enhanced. This demonstrates that our data and training pipeline are not only effective for teaching dynamic interaction but also serve as a powerful method for improving general-purpose geometric intelligence in MLLMs.

\subsection{Comparative Analysis with Visual Sketchpad}

While Visual Sketchpad pioneers the concept of generating visual outputs during reasoning, GeoSketch represents a significant architectural leap forward by shifting from an imperative, code-based framework to a declarative, logic-based one. Visual Sketchpad frames the problem as a code-to-code task, requiring the model to generate executable scripts (e.g., python code). This approach conflates the core challenge of mathematical reasoning with the complex, error-prone task of code generation and relies on manually refined inputs, which fundamentally limits its autonomy and generalizability. In stark contrast, GeoSketch decoules reasoning from rendering. Our model performs pure natural language reasoning to produce geometric inferences. The outcome of this reasoning is either the final solution or a decision to draw an auxiliary line to advance the proof. In the latter case, this decision is encoded as a concise JSON command that instructs our Sketch Action module to modify the image. This allows our model to focus solely on mathematical inference, creating a fully autonomous, end-to-end system that can generalize to any geometry diagram-text pair without human intervention.

The superiority of our declarative approach is quantitatively evident in its efficiency. By abstracting away verbose code, GeoSketch drastically reduces communication overhead. We cut the average input token count from 12899.12 in Visual Sketchpad to just 5692.71, and the average output token count from 1055.75 to 901.46. This significant reduction in token usage not only lowers computational costs and latency but also establishes GeoSketch as a more scalable and practical solution for iterative visual reasoning.

\section{Conclusion}
We introduced GeoSketch, a neural-symbolic framework that reformulates geometric problem-solving as an interactive perception-reasoning-action loop. By enabling Multimodal Large Language Models to dynamically construct auxiliary lines and perform affine transformations, our approach advances beyond static image interpretation to an active, manipulative reasoning process that closely mirrors human expert cognition. Our key contributions include the GeoSketch architecture, which integrates visual perception, symbolic deduction, and executable actions in a closed loop; the GeoSketch Benchmark, a curated set of challenging problems requiring dynamic manipulation; and a two-stage training methodology that uses supervised fine-tuning and reinforcement learning to build a robust and strategic agent.
Experiments demonstrate that integrating MLLMs into the GeoSketch framework yields substantial improvements in problem-solving accuracy over static, direct-evaluation methods. Moreover, our training process enhances the model's fundamental geometric knowledge, leading to better performance even on standard, non-interactive benchmarks. By abstracting diagrams into declarative logic forms, GeoSketch offers a more automated, efficient, and reliable alternative to code-generation-based approaches.

\section*{Reproducibility Statement}

To support reproducibility, we will release the GeoSketch Benchmark (390 problems) and the GeoSketch Fine-Tuning Dataset (2,000 trajectories) in structured formats. The code for the GeoSketch framework, including perception, reasoning, and sketch action modules, will be open-sourced. Pre-trained model weights for GeoSketch-Qwen2.5-VL-7B-SFT and -RL will also be made available. Dataset construction details are in Appendix A; hyperparameters and training configurations are in Appendix C; prompt templates are in Appendix D. Experiments were run on 4×A800 GPUs with CUDA 12.4, and deployment settings are documented for consistency.

\section*{Limitations}

One limitation of this work is that our evaluation focuses solely on final answer accuracy and does not assess the correctness or validity of intermediate reasoning steps. Future work will include step-wise verification to ensure logical soundness throughout the dynamic reasoning process.

\bibliography{iclr2026_conference}
\bibliographystyle{iclr2026_conference}

\clearpage

\appendix

\section{Data Statistics}

\label{app:statistics}
\paragraph{Data Statistics}
As shown in Table~\ref{tab:statistics}, the final answer is explicitly stored as a definitive numerical value (e.g., ``30" for an angle measure in degrees, ``5" for a segment length in specified units), or a precise geometric descriptor (e.g., ``isosceles right triangle" for classifying a figure, ``perpendicular" for describing a line relationship), or a fixed ratio (e.g., ``2:1" for comparing areas of similar triangles, ``1/3" for a segment division proportion).
The most common answer type is numerical value, which is 201 in total. Apart from that, we have 108 problems require the determination of a ratio. Lastly, there are 81 problems that need to answer in the form of a geometric description.

\begin{table}[]
    \centering
    \begin{tabular}{c|c|c|c}
    \hline
      \textbf{Type} & Numerical  &  Ratio & Descriptor \\
    \hline
       \textbf{Count} & 201 & 108 & 81 \\
    \hline
    \end{tabular}
    \caption{Data statistics for the GeoSketch benchmark.}
    \label{tab:statistics}
\end{table}

\paragraph{Multimodal Data Format}
Each problem within the GeoSketch benchmark is formalized into a structured, multimodal data format designed to support the evaluation of dynamic reasoning agents. 
The problem text provides a natural language description of the geometric scenario and the specific question to be answered. It is paired with the initial diagram, an image file that visually presents the given geometric figure without the solution elements. Notably, the image is indispensable to the text; the image cannot be fully inferred or reconstructed solely from the information in the text. Additionally, to bridge the gap between visual perception and symbolic reasoning, we offer an optional formal representation of the initial diagram. This representation takes the form of a structured logical format that encodes geometric objects, their properties, and their interrelationships in a machine-readable way—all derived from our proposed framework.
The final answer is explicitly stored as a definitive numerical value, or a precise geometric descriptor, or a fixed ratio of relations.
Crucially, the benchmark provides a gold solution trajectory. This is a complete, verifiable sequence of reasoning steps that outlines the path from the given state to the solution. Each step in this trajectory pairs a symbolic inference or proof step with the corresponding visual manipulation action, such as drawing an auxiliary line or performing a transformation. This structured format provides a comprehensive ground truth for supervising model behavior and conducting rigorous evaluation. As far as we know, we are the first to provide step-by-step reasoning process verification to dynamic visual geometric problems.

\section{Image Auto Correction}
\label{app:auto-correct}

\begin{figure}
    \centering
    \includegraphics[width=0.7\linewidth]{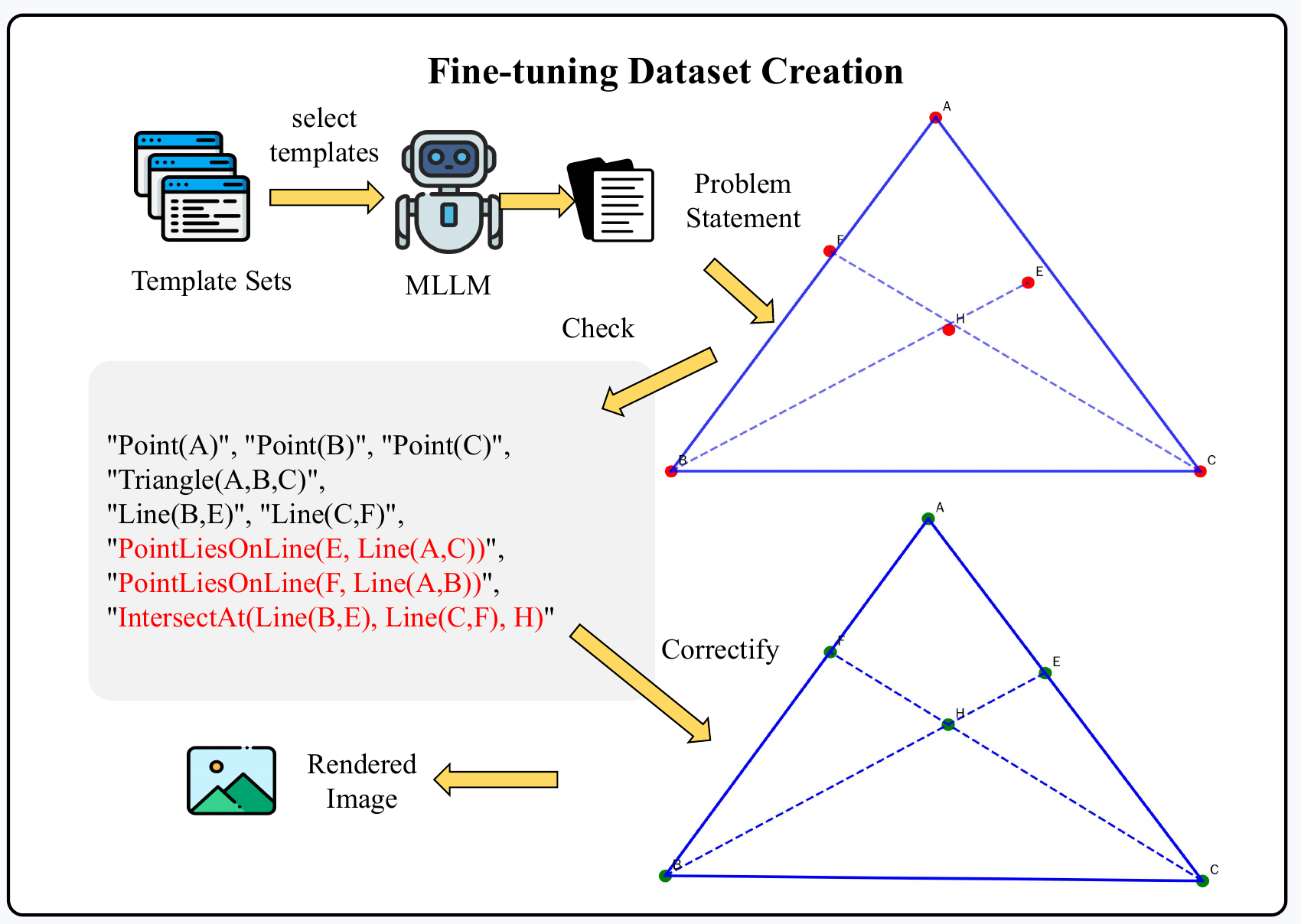}
    \caption{The auto correction mechanism in image generation.}
    \label{fig:auto-correct}
\end{figure}

We build a solid geometry constraint solver(GCS) to parse the geometry relationship among the geometry elements. If the points and lines do not match the relationship, the GCS will get a positive error. Then we utilize the L-BFGS algorithm to optimize the error to zero, then the flawful original geometry diagram become concise, as illustrated in Figure \ref{fig:auto-correct}.

\section{Hyperparameter}
For the SFT stage training, we freeze the vit module and only fine-tune the projector and llm module, the learning rate is set to 1e-5 for both modules. And for RL stage training, we keep the vit frozen and set the learning rate to 1e-6 for both the llm module and the projector. And the experiments were conducted on 4 * A800 GPUs, with 12.4.

For deployment, we maintain the same hardware and software version as the training environment. We use vLLM 0.10.2 to deploy our model, the sampling temperature is set to 0 and the random seed is 42.

\end{document}

%% file: math_commands.tex
\usepackage{amsmath,amsfonts,bm}

\def\eqref#1{equation~\ref{#1}}
\def\1{\bm{1}}

\DeclareMathAlphabet{\mathsfit}{\encodingdefault}{\sfdefault}{m}{sl}
\SetMathAlphabet{\mathsfit}{bold}{\encodingdefault}{\sfdefault}{bx}{n}

%% file: iclr2026_conference.bbl
\begin{thebibliography}{38}
\providecommand{\natexlab}[1]{#1}
\providecommand{\url}[1]{\texttt{#1}}
\expandafter\ifx\csname urlstyle\endcsname\relax
  \providecommand{\doi}[1]{doi: #1}\else
  \providecommand{\doi}{doi: \begingroup \urlstyle{rm}\Url}\fi

\bibitem[Cai et~al.(2024)Cai, Bao, Guo, Zhang, Song, and Zheng]{cai-etal-2024-geogpt4v}
Shihao Cai, Keqin Bao, Hangyu Guo, Jizhi Zhang, Jun Song, and Bo~Zheng.
\newblock {G}eo{GPT}4{V}: Towards geometric multi-modal large language models with geometric image generation.
\newblock In Yaser Al-Onaizan, Mohit Bansal, and Yun-Nung Chen (eds.), \emph{Proceedings of the 2024 Conference on Empirical Methods in Natural Language Processing}, pp.\  750--766, Miami, Florida, USA, November 2024. Association for Computational Linguistics.
\newblock \doi{10.18653/v1/2024.emnlp-main.44}.
\newblock URL \url{https://aclanthology.org/2024.emnlp-main.44/}.

\bibitem[Chen et~al.(2021)Chen, Tang, Qin, Liang, Liu, Xing, and Lin]{DBLP:journals/corr/abs-2105-14517}
Jiaqi Chen, Jianheng Tang, Jinghui Qin, Xiaodan Liang, Lingbo Liu, Eric~P. Xing, and Liang Lin.
\newblock Geoqa: {A} geometric question answering benchmark towards multimodal numerical reasoning.
\newblock \emph{CoRR}, abs/2105.14517, 2021.
\newblock URL \url{https://arxiv.org/abs/2105.14517}.

\bibitem[Chen et~al.(2022)Chen, Li, Qin, Lu, Lin, Chen, and Liang]{chen-etal-2022-unigeo}
Jiaqi Chen, Tong Li, Jinghui Qin, Pan Lu, Liang Lin, Chongyu Chen, and Xiaodan Liang.
\newblock {U}ni{G}eo: Unifying geometry logical reasoning via reformulating mathematical expression.
\newblock In Yoav Goldberg, Zornitsa Kozareva, and Yue Zhang (eds.), \emph{Proceedings of the 2022 Conference on Empirical Methods in Natural Language Processing}, pp.\  3313--3323, Abu Dhabi, United Arab Emirates, December 2022. Association for Computational Linguistics.
\newblock \doi{10.18653/v1/2022.emnlp-main.218}.
\newblock URL \url{https://aclanthology.org/2022.emnlp-main.218/}.

\bibitem[Chen et~al.(2025)Chen, Lee, and Liang]{chen2025interactivesketchpadmultimodaltutoring}
Steven-Shine Chen, Jimin Lee, and Paul~Pu Liang.
\newblock Interactive sketchpad: A multimodal tutoring system for collaborative, visual problem-solving, 2025.
\newblock URL \url{https://arxiv.org/abs/2503.16434}.

\bibitem[Cheng et~al.(2025)Cheng, Chen, Zhang, Fei, Feng, Che, Li, and Qin]{10.1609/aaai.v39i22.34538}
Zihui Cheng, Qiguang Chen, Jin Zhang, Hao Fei, Xiaocheng Feng, Wanxiang Che, Min Li, and Libo Qin.
\newblock Comt: a novel benchmark for chain of multi-modal thought on large vision-language models.
\newblock In \emph{Proceedings of the Thirty-Ninth AAAI Conference on Artificial Intelligence and Thirty-Seventh Conference on Innovative Applications of Artificial Intelligence and Fifteenth Symposium on Educational Advances in Artificial Intelligence}, AAAI'25/IAAI'25/EAAI'25. AAAI Press, 2025.
\newblock ISBN 978-1-57735-897-8.
\newblock \doi{10.1609/aaai.v39i22.34538}.
\newblock URL \url{https://doi.org/10.1609/aaai.v39i22.34538}.

\bibitem[Cho et~al.(2025)Cho, Qin, Liu, Choi, Lee, and Kim]{cho2025planegeometryproblemsolving}
Seunghyuk Cho, Zhenyue Qin, Yang Liu, Youngbin Choi, Seungbeom Lee, and Dongwoo Kim.
\newblock Plane geometry problem solving with multi-modal reasoning: A survey, 2025.
\newblock URL \url{https://arxiv.org/abs/2505.14340}.

\bibitem[Christou et~al.(2005)Christou, Mousoulides, Pittalis, and Pitta-Pantazi]{christou2005problem}
Constantinos Christou, Nicholas Mousoulides, Marios Pittalis, and Demetra Pitta-Pantazi.
\newblock Problem solving and problem posing in a dynamic geometry environment.
\newblock \emph{The Mathematics Enthusiast}, 2\penalty0 (2):\penalty0 125--143, 2005.

\bibitem[Comanici et~al.(2025)Comanici, Bieber, Schaekermann, Pasupat, Sachdeva, Dhillon, Blistein, Ram, Zhang, Rosen, et~al.]{comanici2025gemini}
Gheorghe Comanici, Eric Bieber, Mike Schaekermann, Ice Pasupat, Noveen Sachdeva, Inderjit Dhillon, Marcel Blistein, Ori Ram, Dan Zhang, Evan Rosen, et~al.
\newblock Gemini 2.5: Pushing the frontier with advanced reasoning, multimodality, long context, and next generation agentic capabilities.
\newblock \emph{arXiv preprint arXiv:2507.06261}, 2025.

\bibitem[Cui et~al.(2025)Cui, Yuan, Wang, Li, Du, and Ding]{cui2025drawthoughtunleashingmultimodal}
Zhiqing Cui, Jiahao Yuan, Hanqing Wang, Yanshu Li, Chenxu Du, and Zhenglong Ding.
\newblock Draw with thought: Unleashing multimodal reasoning for scientific diagram generation, 2025.
\newblock URL \url{https://arxiv.org/abs/2504.09479}.

\bibitem[DeepSeek-AI(2025)]{deepseekai2025deepseekr1incentivizingreasoningcapability}
DeepSeek-AI.
\newblock Deepseek-r1: Incentivizing reasoning capability in llms via reinforcement learning, 2025.
\newblock URL \url{https://arxiv.org/abs/2501.12948}.

\bibitem[Fang et~al.(2025)Fang, Duan, Wang, Huang, Li, Yan, Tian, Zeng, Zhao, Dai, Liu, and Li]{fang2025got}
Rongyao Fang, Chengqi Duan, Kun Wang, Linjiang Huang, Hao Li, Shilin Yan, Hao Tian, Xingyu Zeng, Rui Zhao, Jifeng Dai, Xihui Liu, and Hongsheng Li.
\newblock Got: Unleashing reasoning capability of multimodal large language model for visual generation and editing.
\newblock \emph{arXiv preprint arXiv:2503.10639}, 2025.

\bibitem[Freksa et~al.(2019)Freksa, Barkowsky, Falomir, and van~de Ven]{freksa2019geometric}
Christian Freksa, Thomas Barkowsky, Zoe Falomir, and Jasper van~de Ven.
\newblock Geometric problem solving with strings and pins.
\newblock \emph{Spatial Cognition \& Computation}, 19\penalty0 (1):\penalty0 46--68, 2019.

\bibitem[Gao et~al.(2025)Gao, Pi, Zhang, Ye, Zhong, Wang, HONG, Han, Xu, Li, and Kong]{gao2025gllava}
Jiahui Gao, Renjie Pi, Jipeng Zhang, Jiacheng Ye, Wanjun Zhong, Yufei Wang, Lanqing HONG, Jianhua Han, Hang Xu, Zhenguo Li, and Lingpeng Kong.
\newblock G-{LL}a{VA}: Solving geometric problem with multi-modal large language model.
\newblock In \emph{The Thirteenth International Conference on Learning Representations}, 2025.
\newblock URL \url{https://openreview.net/forum?id=px1674Wp3C}.

\bibitem[Guo et~al.(2025)Guo, Chu, Yang, Mo, Shen, Li, et~al.]{guo2025rbenchv}
Meng-Hao Guo, Xuanyu Chu, Qianrui Yang, Zhe-Han Mo, Yiqing Shen, Pei-Lin Li, et~al.
\newblock Rbench-v: A primary assessment for visual reasoning models with multi-modal outputs.
\newblock 2025.
\newblock URL \url{https://arxiv.org/abs/2505.16770}.

\bibitem[He et~al.(2025)He, Lyu, Chen, Guo, and Fung]{he2025matpbenchmllmgoodautomated}
Zhitao He, Zongwei Lyu, Dazhong Chen, Dadi Guo, and Yi~R. Fung.
\newblock Matp-bench: Can mllm be a good automated theorem prover for multimodal problems?, 2025.
\newblock URL \url{https://arxiv.org/abs/2506.06034}.

\bibitem[Hong et~al.(2025)Hong, Yu, Gu, Wang, Gan, Tang, Cheng, Qi, Ji, Pan, et~al.]{hong2025glm}
Wenyi Hong, Wenmeng Yu, Xiaotao Gu, Guo Wang, Guobing Gan, Haomiao Tang, Jiale Cheng, Ji~Qi, Junhui Ji, Lihang Pan, et~al.
\newblock Glm-4.1 v-thinking: Towards versatile multimodal reasoning with scalable reinforcement learning.
\newblock \emph{arXiv e-prints}, pp.\  arXiv--2507, 2025.

\bibitem[Hu et~al.(2025)Hu, Shi, Fu, Roth, Ostendorf, Zettlemoyer, Smith, and Krishna]{10.5555/3737916.3742339}
Yushi Hu, Weijia Shi, Xingyu Fu, Dan Roth, Mari Ostendorf, Luke Zettlemoyer, Noah~A. Smith, and Ranjay Krishna.
\newblock Visual sketchpad: sketching as a visual chain of thought for multimodal language models.
\newblock In \emph{Proceedings of the 38th International Conference on Neural Information Processing Systems}, NIPS '24, Red Hook, NY, USA, 2025. Curran Associates Inc.
\newblock ISBN 9798331314385.

\bibitem[Khanam \& Hussain(2024)Khanam and Hussain]{khanam2024yolov11overviewkeyarchitectural}
Rahima Khanam and Muhammad Hussain.
\newblock Yolov11: An overview of the key architectural enhancements, 2024.
\newblock URL \url{https://arxiv.org/abs/2410.17725}.

\bibitem[Kuang et~al.(2025)Kuang, Shen, Xie, Luo, Xu, Li, Li, Cheng, Lin, and Han]{10.1145/3711680}
Jiayi Kuang, Ying Shen, Jingyou Xie, Haohao Luo, Zhe Xu, Ronghao Li, Yinghui Li, Xianfeng Cheng, Xika Lin, and Yu~Han.
\newblock Natural language understanding and inference with mllm in visual question answering: A survey.
\newblock \emph{ACM Comput. Surv.}, 57\penalty0 (8), March 2025.
\newblock ISSN 0360-0300.
\newblock \doi{10.1145/3711680}.
\newblock URL \url{https://doi.org/10.1145/3711680}.

\bibitem[Lu et~al.(2021)Lu, Gong, Jiang, Qiu, Huang, Liang, and Zhu]{lu-etal-2021-inter}
Pan Lu, Ran Gong, Shibiao Jiang, Liang Qiu, Siyuan Huang, Xiaodan Liang, and Song-Chun Zhu.
\newblock {I}nter-{GPS}: Interpretable geometry problem solving with formal language and symbolic reasoning.
\newblock In Chengqing Zong, Fei Xia, Wenjie Li, and Roberto Navigli (eds.), \emph{Proceedings of the 59th Annual Meeting of the Association for Computational Linguistics and the 11th International Joint Conference on Natural Language Processing (Volume 1: Long Papers)}, pp.\  6774--6786, Online, August 2021. Association for Computational Linguistics.
\newblock \doi{10.18653/v1/2021.acl-long.528}.
\newblock URL \url{https://aclanthology.org/2021.acl-long.528/}.

\bibitem[Lu et~al.(2024)Lu, Bansal, Xia, Liu, Li, Hajishirzi, Cheng, Chang, Galley, and Gao]{lu2024mathvista}
Pan Lu, Hritik Bansal, Tony Xia, Jiacheng Liu, Chunyuan Li, Hannaneh Hajishirzi, Hao Cheng, Kai-Wei Chang, Michel Galley, and Jianfeng Gao.
\newblock Mathvista: Evaluating mathematical reasoning of foundation models in visual contexts.
\newblock In \emph{International Conference on Learning Representations (ICLR)}, 2024.

\bibitem[Ma et~al.(2024)Ma, Zhan, Wong, Li, Sun, Chan, and Chao]{ma2024visaidmathbenchmarkingvisualaidedmathematical}
Jingkun Ma, Runzhe Zhan, Derek~F. Wong, Yang Li, Di~Sun, Hou~Pong Chan, and Lidia~S. Chao.
\newblock Visaidmath: Benchmarking visual-aided mathematical reasoning, 2024.
\newblock URL \url{https://arxiv.org/abs/2410.22995}.

\bibitem[OpenAI(2024)]{openai2024gpt4ocard}
OpenAI.
\newblock Gpt-4o system card, 2024.
\newblock URL \url{https://arxiv.org/abs/2410.21276}.

\bibitem[Peng et~al.(2023)Peng, Fu, Liang, Gao, and Tang]{peng2023geodrl}
Shuai Peng, Di~Fu, Yijun Liang, Liangcai Gao, and Zhi Tang.
\newblock Geodrl: A self-learning framework for geometry problem solving using reinforcement learning in deductive reasoning.
\newblock In \emph{Findings of the Association for Computational Linguistics: ACL 2023}, pp.\  13468--13480, 2023.

\bibitem[Qiao et~al.(2024)Qiao, Tan, Dong, Wu, Sun, Song, GongQue, Lei, Wei, Zhang, Qiao, Zhang, Zong, Xu, Diao, Bao, Li, and Zhang]{qiao2024wemathdoeslargemultimodal}
Runqi Qiao, Qiuna Tan, Guanting Dong, Minhui Wu, Chong Sun, Xiaoshuai Song, Zhuoma GongQue, Shanglin Lei, Zhe Wei, Miaoxuan Zhang, Runfeng Qiao, Yifan Zhang, Xiao Zong, Yida Xu, Muxi Diao, Zhimin Bao, Chen Li, and Honggang Zhang.
\newblock We-math: Does your large multimodal model achieve human-like mathematical reasoning?, 2024.
\newblock URL \url{https://arxiv.org/abs/2407.01284}.

\bibitem[Redmon et~al.(2015)Redmon, Divvala, Girshick, and Farhadi]{Redmon2015YouOL}
Joseph Redmon, Santosh~Kumar Divvala, Ross~B. Girshick, and Ali Farhadi.
\newblock You only look once: Unified, real-time object detection.
\newblock \emph{2016 IEEE Conference on Computer Vision and Pattern Recognition (CVPR)}, pp.\  779--788, 2015.
\newblock URL \url{https://api.semanticscholar.org/CorpusID:206594738}.

\bibitem[Ronneberger et~al.(2015)Ronneberger, Fischer, and Brox]{DBLP:journals/corr/RonnebergerFB15}
Olaf Ronneberger, Philipp Fischer, and Thomas Brox.
\newblock U-net: Convolutional networks for biomedical image segmentation.
\newblock \emph{CoRR}, abs/1505.04597, 2015.
\newblock URL \url{http://arxiv.org/abs/1505.04597}.

\bibitem[Seed et~al.(2025)Seed, Chen, Fan, Liu, Liu, Lin, Wang, Wang, Wei, Xu, et~al.]{seed2025seed1}
ByteDance Seed, Jiaze Chen, Tiantian Fan, Xin Liu, Lingjun Liu, Zhiqi Lin, Mingxuan Wang, Chengyi Wang, Xiangpeng Wei, Wenyuan Xu, et~al.
\newblock Seed1. 5-thinking: Advancing superb reasoning models with reinforcement learning.
\newblock \emph{arXiv preprint arXiv:2504.13914}, 2025.

\bibitem[Shao et~al.(2024)Shao, Wang, Zhu, Xu, Song, Bi, Zhang, Zhang, Li, Wu, and Guo]{shao2024deepseekmath}
Zhihong Shao, Peiyi Wang, Qihao Zhu, Runxin Xu, Junxiao Song, Xiao Bi, Haowei Zhang, Mingchuan Zhang, Y.~K. Li, Y.~Wu, and Daya Guo.
\newblock Deepseekmath: Pushing the limits of mathematical reasoning in open language models, 2024.

\bibitem[Su et~al.(2025)Su, Xia, Guo, Liu, Ma, Qu, Liu, Li, Zeng, Yang, Li, Cheng, Ji, He, and Fung]{su2025thinkingimagesmultimodalreasoning}
Zhaochen Su, Peng Xia, Hangyu Guo, Zhenhua Liu, Yan Ma, Xiaoye Qu, Jiaqi Liu, Yanshu Li, Kaide Zeng, Zhengyuan Yang, Linjie Li, Yu~Cheng, Heng Ji, Junxian He, and Yi~R. Fung.
\newblock Thinking with images for multimodal reasoning: Foundations, methods, and future frontiers, 2025.
\newblock URL \url{https://arxiv.org/abs/2506.23918}.

\bibitem[Trinh et~al.(2024)Trinh, Wu, Le, He, and Luong]{Trinh2024SolvingOG}
Trieu~H. Trinh, Yuhuai Wu, Quoc~V. Le, He~He, and Thang Luong.
\newblock Solving olympiad geometry without human demonstrations.
\newblock \emph{Nature}, 625:\penalty0 476 -- 482, 2024.
\newblock URL \url{https://api.semanticscholar.org/CorpusID:267032902}.

\bibitem[Vinker et~al.(2024)Vinker, Shaham, Zheng, Zhao, Fan, and Torralba]{vinker2024sketchagentlanguagedrivensequentialsketch}
Yael Vinker, Tamar~Rott Shaham, Kristine Zheng, Alex Zhao, Judith~E Fan, and Antonio Torralba.
\newblock Sketchagent: Language-driven sequential sketch generation, 2024.
\newblock URL \url{https://arxiv.org/abs/2411.17673}.

\bibitem[Wang et~al.(2025{\natexlab{a}})Wang, Li, Yin, Ran, and Liu]{wang2025mv}
Peijie Wang, Zhong-Zhi Li, Fei Yin, Dekang Ran, and Cheng-Lin Liu.
\newblock Mv-math: Evaluating multimodal math reasoning in multi-visual contexts.
\newblock In \emph{Proceedings of the Computer Vision and Pattern Recognition Conference}, pp.\  19541--19551, 2025{\natexlab{a}}.

\bibitem[Wang et~al.(2025{\natexlab{b}})Wang, Wu, Zhang, Yan, Liu, Luo, and Fei]{wang2025multimodal}
Yaoting Wang, Shengqiong Wu, Yuecheng Zhang, Shuicheng Yan, Ziwei Liu, Jiebo Luo, and Hao Fei.
\newblock Multimodal chain-of-thought reasoning: A comprehensive survey.
\newblock \emph{arXiv preprint arXiv:2503.12605}, 2025{\natexlab{b}}.

\bibitem[Xu et~al.(2025)Xu, Wang, Wang, Chen, Zhou, Yang, Lu, Li, Wang, Zhu, et~al.]{xu2025visulogic}
Weiye Xu, Jiahao Wang, Weiyun Wang, Zhe Chen, Wengang Zhou, Aijun Yang, Lewei Lu, Houqiang Li, Xiaohua Wang, Xizhou Zhu, et~al.
\newblock Visulogic: A benchmark for evaluating visual reasoning in multi-modal large language models.
\newblock \emph{arXiv preprint arXiv:2504.15279}, 2025.

\bibitem[Yan et~al.(2025)Yan, Su, He, Fu, Zheng, Lyu, Wang, Wang, Wen, and Hu]{yan2025surveymathematicalreasoningera}
Yibo Yan, Jiamin Su, Jianxiang He, Fangteng Fu, Xu~Zheng, Yuanhuiyi Lyu, Kun Wang, Shen Wang, Qingsong Wen, and Xuming Hu.
\newblock A survey of mathematical reasoning in the era of multimodal large language model: Benchmark, method and challenges, 2025.
\newblock URL \url{https://arxiv.org/abs/2412.11936}.

\bibitem[Yang et~al.(2025)Yang, Li, Yang, Zhang, et~al.]{yang2025qwen3technicalreport}
An~Yang, Anfeng Li, Baosong Yang, Beichen Zhang, et~al.
\newblock Qwen3 technical report, 2025.
\newblock URL \url{https://arxiv.org/abs/2505.09388}.

\bibitem[Zhang et~al.(2023)Zhang, Yin, and Liu]{10.24963/ijcai.2023/376}
Ming-Liang Zhang, Fei Yin, and Cheng-Lin Liu.
\newblock A multi-modal neural geometric solver with textual clauses parsed from diagram.
\newblock In \emph{Proceedings of the Thirty-Second International Joint Conference on Artificial Intelligence}, IJCAI '23, 2023.
\newblock ISBN 978-1-956792-03-4.
\newblock \doi{10.24963/ijcai.2023/376}.
\newblock URL \url{https://doi.org/10.24963/ijcai.2023/376}.

\end{thebibliography}
